\documentclass[11pt,a4paper]{article}
\usepackage[hyperref]{acl2017}
\usepackage{times}
\usepackage{latexsym}
\usepackage{url}
\usepackage{graphicx}
\usepackage{amsmath}
\usepackage{multirow}
\graphicspath{ {Figure/} }
\usepackage{tabularx}
\usepackage{subcaption}
\usepackage{xcolor}
\usepackage{pifont}

\aclfinalcopy 

\newcommand*{\affaddr}[1]{#1} 
\newcommand*{\affmark}[1][*]{\textsuperscript{#1}}
\newcommand*{\email}[1]{\texttt{#1}}

\title{Neural-based Natural Language Generation in Dialogue \\using RNN
          Encoder-Decoder with Semantic Aggregation}

\author{%
Van-Khanh Tran\affmark[1,2] and Le-Minh Nguyen\affmark[1]\\
\affaddr{\affmark[1]Japan Advanced Institute of Science and Technology, JAIST\\
					1-1 Asahidai, Nomi, Ishikawa, 923-1292, Japan}\\
\email{\{tvkhanh, nguyenml\}@jaist.ac.jp}\\
\affaddr{\affmark[2]University of Information and Communication Technology, ICTU\\
	Thai Nguyen University, Vietnam}\\
\email{tvkhanh@ictu.edu.vn}%
}

\date{} 

\begin{document}

\maketitle

\begin{abstract}
Natural language generation (NLG) is an important component in spoken dialogue systems. This paper presents a model called Encoder-Aggregator-Decoder which is an extension of an Recurrent Neural Network based Encoder-Decoder architecture.
The proposed Semantic Aggregator consists of two components: an Aligner and a Refiner. The Aligner is a conventional attention calculated over the encoded input information, while the Refiner is another attention or gating mechanism stacked over the attentive Aligner in order to further select and aggregate the semantic elements. 
The proposed model can be jointly trained both sentence planning and surface realization to produce natural language utterances.
The model was extensively assessed on four different NLG domains, in which the experimental results showed that the proposed generator consistently outperforms the previous methods on all the NLG domains.
\end{abstract}

\section{Introduction}\label{sec:introduction}
\begin{table*}[ht]
\centering
\caption{Order issue in natural language generation, in which an incorrect generated sentence has \textcolor{red}{\underline{wrong ordered slots}}.}
\label{tab:issues}
\resizebox{\textwidth}{!}{%
\begin{tabularx}{1.2\textwidth}{lX}
\textbf{Input DA} & \textbf{Compare}(name=\textbf{\textit{Triton 52}}; ecorating=\textbf{\textit{A+}}; family=\textbf{\textit{L7}}; name=\textbf{\textit{Hades 76}}; ecorating=\textbf{\textit{C}}; family=\textbf{\textit{L9}})
 \\
\textbf{INCORRECT} & The \textbf{\textit{Triton 52}} has an \textbf{\textit{A}}+ eco rating and is in the \textbf{\textit{\textcolor{red}{\underline{L9}}}} product family, the \textbf{\textit{Hades 76}} is in the \textbf{\textit{\textcolor{red}{\underline{L7}}}} product family and has a \textbf{\textit{C}} eco rating. \\
\textbf{CORRECT} & The \textbf{\textit{Triton 52}} is in the \textbf{\textit{L7}} product family and has an \textbf{\textit{A+}} eco rating, the \textbf{\textit{Hades 76}} is in the \textbf{\textit{L9}} product family and has a \textbf{\textit{C}} eco rating. \\
\end{tabularx}%
}
\end{table*}

Natural Language Generation (NLG) plays a critical role in a Spoken Dialogue System (SDS), and its task is to convert a meaning representation produced by the dialogue manager into natural language sentences. Conventional approaches to NLG follow a \textit{pipeline} which typically breaks down the task into \textit{sentence planning} and \textit{surface realization}. \textit{Sentence planning} decides the order and structure of a sentence, which is followed by a \textit{surface realization} which converts the sentence structure into final utterance. Previous approaches to NLG still rely on extensive hand-tuning templates and rules that require expert knowledge of linguistic representation. There are some common and widely used approaches to solve NLG problems, including rule-based \cite{cheyer2014method}, corpus-based n-gram generator \cite{oh2000stochastic}, and a trainable generator \cite{Ratnaparkhi:2000:TMS:974305.974331}. 

Recurrent Neural Network (RNN)-based approaches have recently shown promising results in NLG tasks. The RNN-based models have been used for NLG as a joint training model \cite{thwsjy15,wensclstm15} and an end-to-end training network \cite{wen2016network}. A recurring problem in such systems requiring annotated corpora for specific dialogue acts\footnote{A combination of an action type and a set of slot-value pairs. E.g. \textit{inform(name='Piperade'; food='Basque').}} (DAs). More recently, the attention-based RNN Encoder-Decoder (AREncDec) approaches \cite{bahdanau2014neural} have been explored to tackle the NLG problems \cite{wentoward,mei2015talk,duvsek2016sequence,duvsek2016context}. The AREncDEc-based models have also shown improved results on various tasks, e.g., image captioning \cite{xu2015show,yang2016review}, machine translation \cite{luong2015effective,wu2016google}.

To ensure that the generated utterance represents the intended meaning of the given DA, the previous RNN-based models were conditioned on a 1-hot vector representation of the DA. \citet{thwsjy15} proposed a Long Short-Term Memory-based (HLSTM) model which introduced a heuristic gate to guarantee that the slot-value pairs were accurately captured during generation. Subsequently, \citet{wensclstm15} proposed a LSTM-based generator (SC-LSTM) which jointly learned the controlling signal and language model. \citet{wentoward} proposed an AREncDec based generator (ENCDEC) which applied attention mechanism on the slot-value pairs. 

Although these RNN-based generators have worked well, however, they still have some drawbacks, and none of these models significantly outperform the others in solving NLG tasks. While the HLSTM cannot handle cases such as the binary slots (i.e., \textit{yes} and \textit{no}) and slots that take \textit{don't\_care} value in which these slots cannot be directly delexicalized, the SCLSTM model is limited to generalize to the unseen domain, and the ENCDEC model has difficulty to prevent undesirable semantic repetitions during generation. 

To address the above issues, we propose a new architecture, \textit{Encoder-Aggregator-Decoder}, an extension of the AREncDec model, in which the proposed Aggregator has two main components: (i) an Aligner which computes the attention over the input sequence, and (ii) a Refiner which are another attention or gating mechanisms to further select and aggregate the semantic elements. The proposed model can learn from unaligned data by jointly training the sentence planning and surface realization to produce natural language sentences. We conduct comprehensive experiments on four NLG domains and find that the proposed method significantly outperforms the previous methods regarding BLEU \cite{papineni2002bleu} and slot error rate ERR scores \cite{wensclstm15}. We also found that our generator can produce high-quality utterances with correctly ordered than those in the previous methods (see Table \ref{tab:issues}). To sum up, we make two key contributions in this paper:
\begin{itemize}
	\item We present a semantic component called \textit{Aggregator} which is easy integrated into existing (attentive) RNN encoder-decoder architecture, resulting in an end-to-end generator that empirically improved performance in comparison with the previous approaches.
	\item We present several different choices of attention and gating mechanisms which can be effectively applied to the proposed semantic Aggregator.
\end{itemize}

In Section \ref{sec:relatedwork}, we review related works. The proposed model is presented in Section \ref{sec:method}. Section \ref{sec:experiments} describes datasets, experimental setups and evaluation metrics. The results and analysis are presented in Section \ref{sec:resultsandanalysis}. We conclude with a brief of summary and future work in Section \ref{sec:conclusion}. 

\section{Related Work}\label{sec:relatedwork}
Conventional approaches to NLG traditionally divide the task into a pipeline of sentence planning and surface realization. The conventional methods still rely on the handcrafted rule-based generators or rerankers. \citet{oh2000stochastic} proposed a class-based n-gram language model (LM) generator which can learn to generate the sentences for a given dialogue act and then select the best sentences using a rule-based reranker. \citet{Ratnaparkhi:2000:TMS:974305.974331} later addressed some of the limitation of the class-based LMs by proposing a method based on a syntactic dependency tree. A phrase-based generator based on factored LMs was introduced by \citet{mairesse2014stochastic}, that can learn from a semantically aligned corpus.

Recently, RNNs-based approaches have shown promising results in the NLG domain. \citet{vinyals2015show,karpathy2015deep} applied RNNs in setting of multi-modal to generate caption for images. \citet{zhang2014chinese} also proposed a generator using RNNs to create Chinese poetry.
For task-oriented dialogue systems, \citet{thwsjy15} combined two TNN-based models with a CNN reranker to generate required utterances. \citet{wensclstm15} proposed SC-LSTM generator which proposed an additional "reading" cell to the traditional LSTM cell to learn the gating mechanism and language model jointly. A recurring problem in such systems lacking of sufficient domain-specific annotated corpora. \citet{wen2016multi} proposed an out-of-domain model which is trained on counterfeited datasets by using semantic similar slots from the target-domain dataset instead of the slots belonging to the out-of-domain dataset. The empirical results indicated that the model can obtain a satisfactory results with a small amount of in-domain data by fine-tuning the target-domain on the out-of-domain trained model. 
\begin{figure}[!ht]
	\centering
    \includegraphics[width=0.36\textwidth, height=6cm]{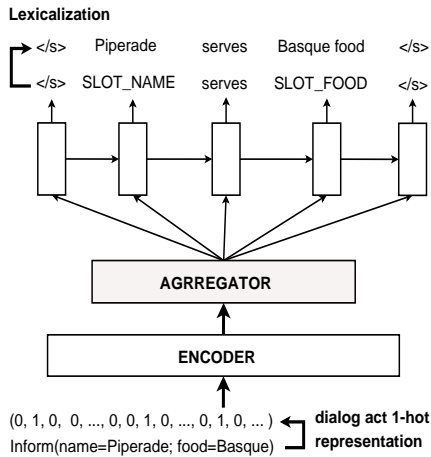}
    \caption{Unfold presentation of the RNN-based neural language generator. The encoder part is subject to various designs, while the decoder is an RNN network.}
    \label{fig:nlg-model}
\end{figure}

More recently, attentional RNN encoder-decoder based models \cite{bahdanau2014neural} have shown improved results in a variety of tasks. \citet{yang2016review} presented a review network in solving the image captioning task, which produces a compact thought vector via reviewing all the input information encoded by the encoder. \citet{mei2015talk} proposed attentional RNN encoder-decoder based model by introducing two layers of attention to model content selection and surface realization. More close to our work, \citet{wentoward} proposed an attentive encoder-decoder based generator, which applied the attention mechanism over the slot-value pairs. The model indicated a domain scalability when a very limited proportion of training data is available.
\section{Recurrent Neural Language Generator}\label{sec:method}

\begin{figure}[!ht]
	\centering
    \includegraphics[width=0.36\textwidth, height=6cm]{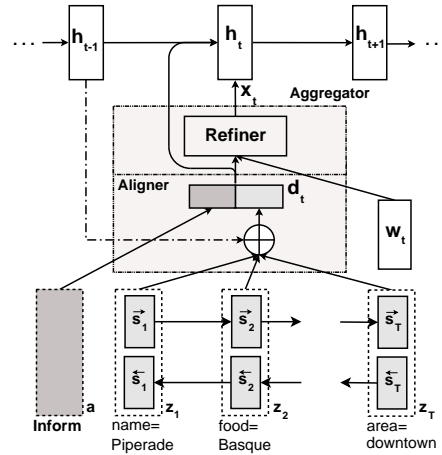}
    \caption{The RNN Encoder-Aggregator-Decoder for NLG proposed in this paper. The output side is an RNN network while the input side is a DA embedding with aggregation mechanism. The Aggregator consists of two parts: an Aligner and a Refiner. The lower part Aligner is an attention over the DA representation calculated by a Bidirectional RNN. Note that the action type embedding $\textbf{a}$ is not included in the attention mechanism since its task is controlling the style of the sentence. The higher part Refiner computes the new input token $\textbf{x}_{t}$ based on the original input token $\textbf{w}_{t}$ and the dialogue act attention $\textbf{d}_{t}$. There are several choices for Refiner, i.e., gating mechanism or attention mechanism.}
    \label{fig:AoAGEN-model}
\end{figure}
The recurrent language generator proposed in this paper is based on a neural net language generator \cite{wentoward} which consists of three components: an encoder to incorporate the target meaning representation as the model inputs, an aggregator to align and control the encoded information, and a decoder to generate output sentences. The generator architecture is shown in Figure \ref{fig:nlg-model}. While the decoder typically uses an RNN model, there is a variety of ways to choose the encoder because it depends on the nature of the meaning representation and the interaction between semantic elements. The encoder first encodes the input meaning representation, then the aggregator with a feature selecting or an attention-based mechanism is used to aggregate and select the input semantic elements. The input to the RNN decoder at each time step is a 1-hot encoding of a token\footnote{Input texts are delexicalized in which slot values are replaced by its corresponding slot tokens.} and the aggregated input vector. The output of RNN decoder represents the probability distribution of the next token given the previous token, the dialogue act representation, and the current hidden state. At generation time, we can sample from this conditional distribution to obtain the next token in a generated sentence, and feed it as the next input to the RNN decoder. This process finishes when a stop sign is generated \cite{karpathy2015deep}, or some constraint is reached \cite{zhang2014chinese}. The network can generate a sequence of tokens which can be lexicalized\footnote{The process in which slot token is replaced by its value.} to form the required utterance. 
\subsection{Gated Recurrent Unit}\label{subsec:gru}
The encoder and decoder of the proposed model use a Gated Recurrent Unit (GRU) network proposed by \citet{bahdanau2014neural}, which maps an input sequence $\textbf{W} = [\textbf{w}_{1}, \textbf{w}_{2}, .., \textbf{w}_{T}]$ to a sequence of states $\textbf{H} = [\textbf{h}_{1}, \textbf{h}_{2}, .., \textbf{h}_{T}]$ as follows:
\begin{equation}\label{eq:r-t-2}
\begin{aligned}
\textbf{r}_{i}&=\sigma(\textbf{W}_{rw}\textbf{w}_{i}+\textbf{W}_{rh}\textbf{h}_{i-1})\\
\textbf{u}_{i}&=\sigma(\textbf{W}_{uw}\textbf{w}_{i}+\textbf{W}_{uh}\textbf{h}_{i-1})\\
\tilde{\textbf{h}_{i}}&=\tanh(\textbf{W}_{hw}\textbf{w}_{i}+\textbf{r}_{i}\odot \textbf{W}_{hh}\textbf{h}_{i-1})\\
\textbf{h}_{i}&= \textbf{u}_{i} \odot \textbf{h}_{i-1} + (1-\textbf{u}_{i}) \odot \tilde{\textbf{h}_{i}}
\end{aligned}
\end{equation}
where: $\odot$ denotes the element-wise multiplication, $\textbf{r}_{i}$ and $\textbf{u}_{i}$ are called the reset and update gates respectively, and $\tilde{\textbf{h}_{i}}$ is the candidate activation.
\subsection{Encoder}
The encoder uses a separate parameterization of the slots and values. It encodes the source information into a distributed vector representation $\textbf{z}_{i}$ which is a concatenation of embedding vector representation of each slot-value pair, and is computed by:
\begin{equation}\label{eq:z-i-1}
\textbf{z}_{i} = \textbf{o}_{i} \oplus \textbf{v}_{i}
\end{equation}
where: $\textbf{o}_{i}$ and $\textbf{v}_{i}$ are the $i$-th slot and value embedding, respectively. The \textit{i} index runs over the given slot-value pairs.
In this study, we use a Bidirectional GRU (Bi-GRU) to encode the sequence of slot-value pairs\footnote{We treat the set of slot-value pairs as a sequence and use the order specified by slot's name (e.g., slot \textit{area} comes first, \textit{price} follows \textit{area}). We have tried treating slot-value pair sequence as natural order as appear in the DA, which even yielded worse results.} embedding. The Bi-GRU consists of forward and backward GRUs. The forward GRU reads the sequence of slot-value pairs from left-to-right and calculates the forward hidden states ($\overrightarrow{s_{1}}, .., \overrightarrow{s_{K}}$). The backward GRU reads the slot-value pairs from right-to-left, resulting in a sequence of backward hidden states ($\overleftarrow{s_{1}}, .., \overleftarrow{s_{K}}$). We then obtain the sequence of hidden states $\textbf{S}=[\textbf{s}_{1}, \textbf{s}_{2}, .., \textbf{s}_{K}]$ where $\textbf{\textbf{s}}_{i}$ is a sum of the forward hidden state $\overrightarrow{s_{i}}$ and the backward one $\overleftarrow{s_{i}}$ as follows: 
\begin{equation}\label{eq:s-i}
\textbf{s}_{i}=\overrightarrow{s_{i}} + \overleftarrow{s_{i}}
\end{equation} 
\subsection{Aggregator}\label{subsec:aggregator}
The Aggregator consists of two components: an Aligner and a Refiner. The Aligner computes the dialogue act representation while the choices for Refiner can be varied. 

Firstly, the Aligner calculates dialogue act embedding $\textbf{d}_{t}$ as follows:
\begin{equation}\label{eq:d-t}
\textbf{d}_{t} = \textbf{a} \oplus \sum\nolimits_{i}\alpha_{t,i} \textbf{s}_{i}
\end{equation} 
where: \textbf{a} is vector embedding of the action type, $\oplus$ is vector concatenation, and $\alpha_{t,i}$ is the weight of \textit{i}-th slot-value pair calculated by the attention mechanism:
\begin{equation}
\begin{aligned}
\alpha_{t,i} &= \frac{\exp(e_{t,i})}{\sum\nolimits_{j}\exp(e_{t,j})}\\
e_{t,i}&=a(\textbf{s}_{i}, \textbf{h}_{t-1})\\
a(\textbf{s}_{i}, \textbf{h}_{t-1}) &= \textbf{v}_{a}^{\top}\tanh(\textbf{W}_{a}\textbf{s}_{i} + \textbf{U}_{a}\textbf{h}_{t-1})
\end{aligned}
\end{equation}
where: $a(.,.)$ is an alignment model,$\textbf{v}_{a}, \textbf{W}_{a}, \textbf{U}_{a}$ are the weight matrices to learn.

Secondly, the Refiner calculates the new input $\textbf{x}_{t}$ based on the original input token $\textbf{w}_{t}$ and the DA representation. There are several choices to formulate the Refiner such as gating mechanism or attention mechanism. For each input token $\textbf{w}_{t}$, the selected mechanism module computes the new input $\textbf{x}_{t}$ based on the dialog act representation $\textbf{d}_{t}$ and the input token embedding $\textbf{w}_{t}$, and is formulated by:
\begin{equation}\label{eq:x-t-0}
\textbf{x}_{t} = f_{R}(\textbf{d}_{t}, \textbf{w}_{t})
\end{equation}
where: $f_{R}$ is a refinement function, in which each input token is refined (or filtered) by the dialogue act attention information before putting into the RNN decoder. By this way, we can represent the whole sentence based on this refined input using RNN model.

\subparagraph*{Attention Mechanism:} Inspired by work of \citet{cui2016attention}, in which an attention-over-attention was introduced in solving reading comprehension tasks, we place another attention applied for Refiner over the attentive Aligner, resulting in a model Attentional Refiner over Attention (ARoA).

\begin{itemize}
	\item ARoA with Vector (\textit{ARoA-V}): We use a simple attention where each input token representation is weighted according to dialogue act attention as follows:
      \begin{equation}\label{eq:beta-t-ARoA-V}
      \begin{aligned}
       \beta_{t}&= \sigma(\textbf{V}_{ra}^{\top} \textbf{d}_{t}) \\
      	f_{R}(\textbf{d}_{t}, \textbf{w}_{t})&=\beta_{t} * \textbf{w}_{t}
      \end{aligned}
      \end{equation}
where: $\textbf{V}_{ra}$ is a refinement attention vector which is used to determine the dialogue act attention strength, and $\sigma$ is sigmoid function to normalize the weight $\beta_{t}$ between $0$ and $1$.
	\item ARoA with Matrix (\textit{ARoA-M}): ARoA-V uses only a vector $\textbf{V}_{ra}$ to weight the DA attention. It may be better to use a matrix to control the attention information. The Equation \ref{eq:beta-t-ARoA-V} is modified as follows:
      \begin{equation}\label{eq:beta-t-ARoA-M}
      \begin{aligned}
      \textbf{V}_{ra}&=\textbf{W}_{aw}\textbf{w}_{t}\\
      \beta_{t}&= \sigma(\textbf{V}_{ra}^{\top} \textbf{d}_{t}) \\
      f_{R}(\textbf{d}_{t}, \textbf{w}_{t})&=\beta_{t} * \textbf{w}_{t}
      \end{aligned}
      \end{equation}
where: $\textbf{W}_{aw}$ is a refinement attention matrix.
\item ARoA with Context (\textit{ARoA-C}): The attention in ARoA-V and ARoA-M may not capture the relationship between multiple tokens. In order to add context information into the attention process, we modify the attention weights in Equation \ref{eq:beta-t-ARoA-M} with additional history information $\textbf{h}_{t-1}$:
      \begin{equation}\label{eq:beta-t}
      \begin{aligned}
  \textbf{V}_{ra}=\textbf{W}_{aw}\textbf{w}_{t}+\textbf{W}_{ah}\textbf{h}_{t-1}\\
      \beta_{t}= \sigma(\textbf{V}_{ra}^{\top} \textbf{d}_{t}) \\
      f_{R}(\textbf{d}_{t}, \textbf{w}_{t}, \textbf{h}_{t-1})=\beta_{t} * \textbf{w}_{t}
      \end{aligned}
       \end{equation}
where: $\textbf{W}_{aw}, \textbf{W}_{ah}$ are parameters to learn, $\textbf{V}_{ra}$ is the refinement attention vector same as above, which contains both DA attention and context information.
\end{itemize} 

\subparagraph*{Gating Mechanism:} We use simple element-wise operators (multiplication or addition) to gate the information between the two vectors $\textbf{d}_{t}$ and $\textbf{w}_{t}$ as follows:
\begin{itemize}
	\item Multiplication (\textit{GR-MUL}): The element-wise multiplication plays a part in word-level matching which learns not only the vector similarity, but also preserve information about the two vectors:
    \begin{equation}\label{eq:beta-t}
	f_{R}(\textbf{d}_{t}, \textbf{w}_{t})= \textbf{W}_{gd}\textbf{d}_{t} \odot \textbf{w}_{t}
	\end{equation}
    \item Addition (\textit{GR-ADD}):
    \begin{equation}\label{eq:beta-t}
	f_{R}(\textbf{d}_{t}, \textbf{w}_{t})= \textbf{W}_{gd}\textbf{d}_{t} + \textbf{w}_{t}
	\end{equation}
\end{itemize}
\subsection{Decoder}\label{subsec:decoder}
The decoder uses a simple GRU model as described in Section \ref{subsec:gru}. In this work, we propose to apply the DA representation and the refined inputs deeper into the GRU cell. Firstly, the GRU reset and update gates can be further influenced on the DA attentive information  $\textbf{d}_{t}$. The reset and update gates are modified as follows:
\begin{equation}\label{eq:r-t} 
\begin{aligned}
\textbf{r}_{t}&=\sigma ({\textbf{W}_{rx}\textbf{x}_{t}+\textbf{W}_{rh}\textbf{h}_{t-1}+\textbf{W}_{rd}\textbf{d}}_{t})
\\
\textbf{u}_{t}&=\sigma (\textbf{W}_{ux}\textbf{x}_{t}+\textbf{W}_{uh}\textbf{h}_{t-1}+\textbf{W}_{ud}\textbf{d}_{t})
\end{aligned}
\end{equation}
where: $\textbf{W}_{rd}$ and $\textbf{W}_{ud}$ act like background detectors that learn to control the style of the generating sentence. By this way, the reset and update gates learn not only the long-term dependency but also the attention information from the dialogue act and the previous hidden state. Secondly, the candidate activation $\tilde{\textbf{h}_{t}}$ is also modified to depend on the DA representation as follows:
\begin{equation}\label{eq:h-t-2}
\begin{split}
\tilde{\textbf{h}_{t}}=\tanh(\textbf{W}_{hx}\textbf{x}_{t}+\textbf{r}_{t}\odot \textbf{W}_{hh}\textbf{h}_{t-1}\\+\textbf{W}_{hd}\textbf{d}_{t})
+ \tanh(\textbf{W}_{dc}\textbf{d}_{t})
\end{split}
\end{equation}
 The hidden state is then computed by:
\begin{equation}
\textbf{h}_{t}= \textbf{u}_{t} \odot \textbf{h}_{t-1} + (1-\textbf{u}_{t}) \odot \tilde{\textbf{h}_{t}}
\end{equation}
Finally, the output distribution is computed by applying a softmax function $g$, and the distribution is sampled to obtain the next token,
\begin{equation}\label{eq:p-t-1}
\begin{split}
P(w_{t+1}\mid w_{t}, w_{t-1},...w_{0},\textbf{z}) = g(\textbf{W}_{ho}\textbf{h}_{t}) \\
w_{t+1}\sim P(w_{t+1}\mid w_{t}, w_{t-1},...w_{0},\textbf{z})
\end{split}
\end{equation}

\subsection{Training}\label{subsec:training}
The objective function was the negative log-likelihood and simply computed by:
\begin{equation}\label{eq:c-f-1}
F(\theta) = -\sum_{t=1}^{T}\textbf{y}_{t}^{\top}\log{\textbf{p}_{t}}
\end{equation}
where: $\textbf{y}_{t}$ is the ground truth word distribution, $\textbf{p}_{t}$ is the predicted word distribution, $T$ is length of the input sequence. 
The proposed generators were trained by treating each sentence as a mini-batch with \textit{$l_{2}$} regularization added to the objective function for every 10 training examples. The pre-trained word vectors \cite{pennington2014glove} were used to initialize the model. The generators were optimized by using stochastic gradient descent and back propagation through time \cite{werbos1990backpropagation}. To prevent over-fitting, we implemented early stopping using a validation set as suggested by \citet{mikolov2010recurrent}.

\begin{table*}[!ht]
\centering
\caption{Comparison performance on four datasets in terms of the BLEU and the error rate ERR(\%) scores; \textbf{bold} denotes the best and \textbf{\textit{italic}} shows the second best model. The results were produced by training each network on 5 random initialization and selected model with the highest validation BLEU score. $^{\sharp}$ denotes the Attention-based Encoder-Decoder model.}
\label{tab:tab-performance}
\scalebox{0.95}{
\begin{tabular}{ccccccccc}
\hline 
\multirow{2}{*}{Model} & \multicolumn{2}{c}{\textbf{Restaurant}} & \multicolumn{2}{c}{\textbf{Hotel}} & \multicolumn{2}{c}{\textbf{Laptop}} & \multicolumn{2}{c}{\textbf{TV}} \\ \cline{2-9} 
 & BLEU & ERR & BLEU & ERR & BLEU & ERR & BLEU & ERR \\ \hline
HLSTM               & 0.7466 & 0.74\% & 0.8504 & 2.67\% & 0.5134 & 1.10\% & 0.5250 & 2.50\% \\
SCLSTM              & 0.7525 & 0.38\% & 0.8482 & 3.07\% & 0.5116 & 0.79\% & 0.5265 & 2.31\% \\
ENCDEC$^{\sharp}$   & 0.7398 & 2.78\% & 0.8549 & 4.69\% & 0.5108 & 4.04\% & 0.5182 & 3.18\% \\ \hline \hline

GR-ADD$^{\sharp}$   & 0.7742 & 0.59\% & 0.8848 & 1.54\% & \textbf{\textit{0.5221}}& \textbf{\textit{0.54}}\% & 0.5348 & 0.77\% \\
GR-MUL$^{\sharp}$   & 0.7697 & 0.47\% & 0.8854 & 1.47\% & 0.5200 & 1.15\% & 0.5349 & 0.65\% \\ \hline
ARoA-V$^{\sharp}$	& 0.7667 & \textbf{\textit{0.32}}\% & 0.8814 & \textbf{0.97}\% & 0.5195 & 0.56\% & \textbf{\textit{0.5369}} & 0.81\% \\ 
ARoA-M$^{\sharp}$	& \textbf{0.7755} & \textbf{0.30}\% & \textbf{0.8920} & \textbf{\textit{1.13}}\% & \textbf{0.5223} & \textbf{0.50}\% & \textbf{0.5394} & \textbf{0.60}\% \\
ARoA-C$^{\sharp}$	& \textbf{\textit{0.7745}} & 0.45\% & \textbf{\textit{0.8878}} & 1.31\% & 0.5201 & 0.88\% & 0.5351 & \textbf{\textit{0.63}}\% \\ 
\end{tabular}
}
\end{table*}

\begin{table*}[!ht]
\centering
\caption{Comparison performance of variety of the proposed models on four dataset in terms of the BLEU and the error rate ERR(\%) scores; \textbf{bold} denotes the best and \textbf{\textit{italic}} shows the second best model. The first two models applied gating mechanism to Refiner component while the last three models used attention over attention mechanism. The results were averaged over 5 randomly initialized networks.}
\label{tab:tab-average-performance}
\scalebox{0.95}{
\begin{tabular}{ccccccccc}
\hline
\multirow{2}{*}{Model} & \multicolumn{2}{c}{\textbf{Restaurant}} & \multicolumn{2}{c}{\textbf{Hotel}} & \multicolumn{2}{c}{\textbf{Laptop}} & \multicolumn{2}{c}{\textbf{TV}} \\ \cline{2-9} 
 & BLEU & ERR & BLEU & ERR & BLEU & ERR & BLEU & ERR \\ \hline
GR-ADD & 0.7685 & 0.63\% & \textbf{\textit{0.8838}} & 1.67\% & \textbf{\textit{0.5194}} & \textbf{\textit{0.66}}\% & \textbf{\textit{0.5344}} & 0.75\% \\
GR-MUL 		& 0.7669 & \textbf{\textit{0.61}}\% & 0.8836 & 1.40\% & 0.5184 & 1.01\% & 0.5328 & 0.73\% \\\hline 
ARoA-V & 0.7673 & 0.62\% & 0.8817 & \textbf{\textit{1.27}}\% & 0.5185 & 0.73\% & 0.5336 & 0.68\% \\ 
ARoA-M & \textbf{0.7712} & \textbf{0.50}\% & \textbf{0.8851} & \textbf{1.14}\% & \textbf{0.5201} & \textbf{0.62}\% & \textbf{0.5350} & \textbf{0.62}\% \\
ARoA-C & \textbf{\textit{0.7690}} & 0.70\% & 0.8835 & 1.44\% & 0.5181 & 0.78\% & 0.5307 & \textbf{\textit{0.64}}\% \\
\end{tabular}
}
\end{table*}

\subsection{Decoding}\label{subsec:decoding}
The decoding consists of two phases: (i) over-generation, and (ii) reranking. In the over-generation, the generator conditioned on the given DA uses a beam search to generate a set of candidate responses. In the reranking, the cost of the generator is computed to form the reranking score $R$ as follows:
\begin{equation}\label{eq:r-score-1}
R = -\sum_{t=1}^{T}\textbf{y}_{t}^{\top}\log{\textbf{p}_{t}} + \lambda ERR
\end{equation}
where $\lambda$ is a trade off constant and is set to a large value in order to severely penalize nonsensical outputs. The slot error rate $ERR$, which is the number of slots generated that is either redundant or missing, and is computed by:
\begin{equation}
ERR = \frac{p + q}{N}
\end{equation}
where: $N$ is the total number of slots in DA, and $p$, $q$ is the number of missing and redundant slots, respectively. Note that the ERR reranking criteria cannot handle arbitrary slot-value pairs such as \textit{binary} slots or slots that take the \textit{don’t\_care} value because these slots cannot be delexicalized and matched.

\section{Experiments}\label{sec:experiments}
We conducted an extensive set of experiments to assess the effectiveness of our model using several metrics, datasets, and model architectures, in order to compare to prior methods.
\subsection{Datasets}\label{subsec:datasets}
We assessed the proposed models using four different NLG domains: finding a restaurant, finding a hotel, buying a laptop, and buying a television. The Restaurant and Hotel were collected in \cite{wensclstm15} which contain around 5K utterances and 200 distinct DAs. The Laptop and TV datasets have been released by \citet{wen2016multi}. These datasets contain about 13K distinct DAs in the Laptop domain and 7K distinct DAs in the TV. Both Laptop and TV datasets have a much larger input space but only one training example for each DA so that the system must learn partial realization of concepts and be able to recombine and apply them to unseen DAs. 
As a result, the NLG tasks for the Laptop and TV datasets become much harder. 

\subsection{Experimental Setups}\label{subsec:experimental-setups}
\begin{figure*}[!ht]
	\centering 
    \includegraphics[width=0.95\textwidth, height=4.5cm]{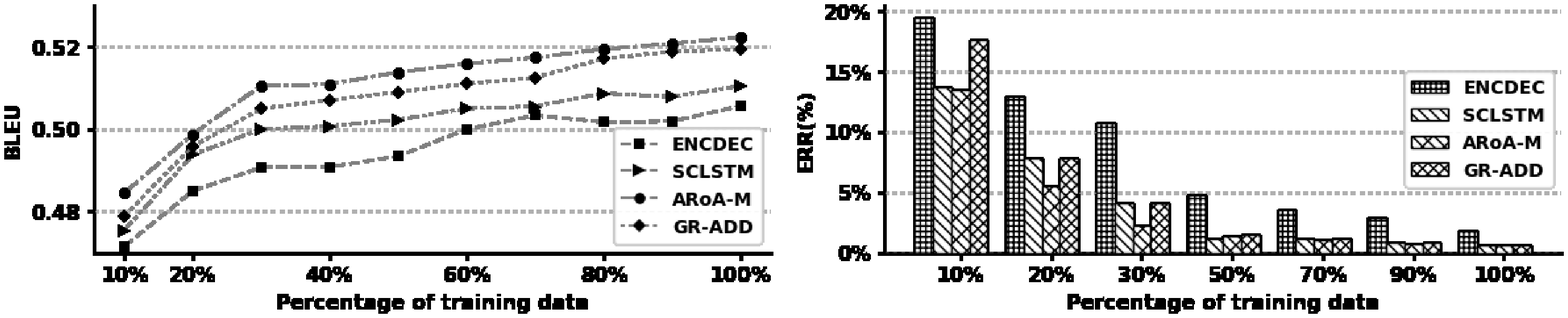}
    \caption{Performance comparison of the four models trained on Laptop (unseen) domain.}
    \label{fig:laptop-performances}
\end{figure*}

\begin{figure*}[!ht]
	\centering 
    \includegraphics[width=0.95\textwidth, height=4cm]{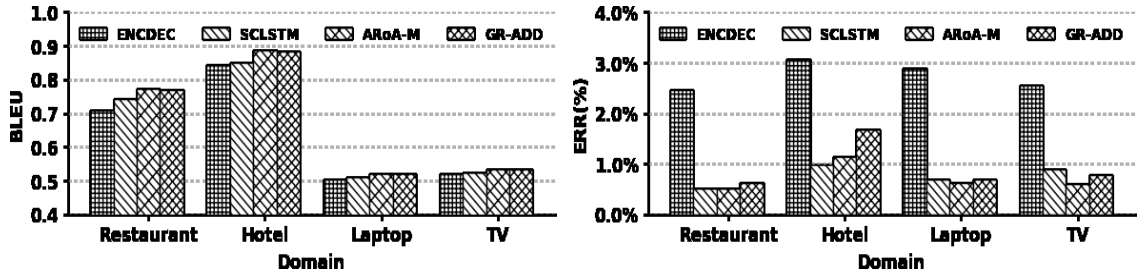}
    \caption{Performance comparison of the general models on four different domains.}
    \label{fig:general-models}
\end{figure*}

The generators were implemented using the TensorFlow library \cite{abadi2016tensorflow} and trained by partitioning each of the datasets into training, validation and testing set in the ratio 3:1:1. The hidden layer size was set to be 80 for all cases, and the generators were trained with a $70\%$ of dropout rate. We perform 5 runs with different random initialization of the network and the training is terminated by using early stopping as described in Section \ref{subsec:training}. We select a model that yields the highest BLEU score on the validation set as shown in Table \ref{tab:tab-performance}. Since the trained models can differ depending on the initialization, we also report the results which were averaged over 5 randomly initialized networks. Note that, except the results reported in Table \ref{tab:tab-performance}, all the results shown were averaged over 5 randomly initialized networks. The decoder procedure used beam search with a beam width of 10. We set $\lambda$ to 1000 to severely discourage the reranker from selecting utterances which contain either redundant or missing slots. For each DA, we over-generated 20 candidate utterances and selected the top 5 realizations after reranking. Moreover, in order to better understand the effectiveness of our proposed methods, we (1) trained the models on the Laptop domain with a varied proportion of training data, starting from $10\%$ to $100\%$ (Figure \ref{fig:laptop-performances}), and (2) trained general models by merging all the data from four domains together and tested them in each individual domain (Figure \ref{fig:general-models})
.
\subsection{Evaluation Metrics and Baselines}\label{subsec:evaluation-metrics}
The generator performance was assessed by using two objective evaluation metrics: the BLEU score and the slot error rate ERR. Both metrics were computed by adopting code from an open source benchmark NLG toolkit\footnotemark. We compared our proposed models against three strong baselines from the open source benchmark toolkit. The results have been recently published as an NLG benchmarks by the Cambridge Dialogue Systems Group\footnotemark[\value{footnote}], including \textit{HLSTM}, \textit{SCLSTM}, and \textit{ENCDEC} models.
\footnotetext{https://github.com/shawnwun/RNNLG}
\begin{figure*}[!ht]
	\centering
    \includegraphics[width=0.9\textwidth, height=4cm]{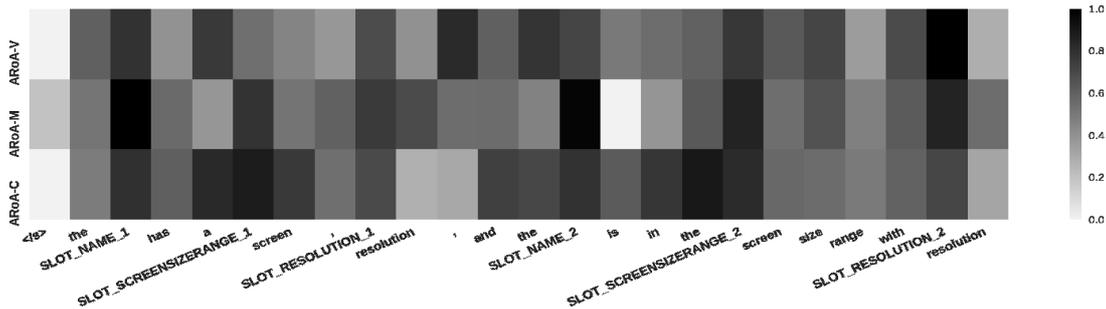}
    \caption{A comparison on attention behavior of three models in a sentence on given \textbf{\textit{DA}} with sequence of slots  [\textit{Name\_1, ScreenSizeRange\_1, Resolution\_1, Name\_2, ScreenSizeRange\_2, Resolution\_2}].}
    \label{fig:attention}
\end{figure*}

\begin{table*}[ht]
\centering
\caption{Comparison of top responses generated for some input dialogue acts between different models. Errors are marked in color (\textcolor{blue}{missing}, \textcolor{red}{misplaced} slot-value pair). \textbf{$^{\dag}$} and \textbf{$^{\natural}$} denotes the baselines and the proposed models, respectively.}
\label{tab:comparison}
\resizebox{\textwidth}{!}{%
\begin{tabularx}{1.2\textwidth}{cX}
\textbf{Model}&\textbf{Generated Responses in Laptop domain} \\ \hline
\textbf{\textit{Input DA}} & \textit{compare(name=`aristaeus 59'; screensizerange=`large'; resolution=`1080p'; name=`charon 61'; screensizerange=`medium'; resolution=`720p')} \\
\textbf{\textit{Reference}} & \textit{Compared to aristaeus 59 which is in the large screen size range and has 1080p resolution, charon 61 is in the medium screen size range and has 720p resolution. Which one do you prefer?} \\

ENCDEC\textbf{$^{\dag}$} & the aristaeus 59 has a large screen , the charon 61 has a medium screen and 1080p resolution [\textcolor{red}{1080p}, \textcolor{blue}{720p}] \\

HLSTM\textbf{$^{\dag}$} & the aristaeus 59 has a large screen size range and has a 1080p resolution and 720p resolution [\textcolor{red}{720p}, \textcolor{blue}{charon 61}, \textcolor{blue}{medium}] \\

SCLSTM\textbf{$^{\dag}$} & the aristaeus 59 has a large screen and 1080p resolution , the charon 61 has a medium screen and 720p resolution \\

GR-ADD\textbf{$^{\natural}$} & the aristaeus 59 has a large screen size and 1080p resolution , the charon 61 has a medium screen size and 720p resolution\\

GR-MUL\textbf{$^{\natural}$} & the aristaeus 59 has a large screen size and 1080p resolution , the charon 61 has a medium screen size and 720p resolution .\\

ARoA-V\textbf{$^{\natural}$} & the aristaeus 59 has a large screen size and 1080p resolution , the charon 61 has a medium screen size , and has a 720p resolution \\

ARoA-M\textbf{$^{\natural}$} & the aristaeus 59 has a large screen and 1080p resolution , the charon 61 has a medium screen and 720p resolution \\ 

ARoA-C\textbf{$^{\natural}$} & the aristaeus 59 has a large screen size and 1080p resolution , the charon 61 has a medium screen size range and 720p resolution \\ \hline
\end{tabularx}%
}
\end{table*}

\section{Results and Analysis}\label{sec:resultsandanalysis}
\subsection{Results}\label{subsec:results}
We conducted extensive experiments on the proposed models with varied setups of Refiner and compared against the previous methods. Overall, the proposed models consistently achieve the better performances regarding both evaluation metrics across all domains. 

Table \ref{tab:tab-performance} shows a comparison between the \textit{AREncDec} based models (the models with $^{\sharp}$ symbol) in which the proposed models significantly reduce the slot error rate across all datasets by a large margin about $2\%$ to $4\%$ that are also improved performances on the BLEU score when comparing the proposed models against the previous approaches. Table \ref{tab:tab-average-performance} further shows the stable strength of our models since the results' pattern stays unchanged compared to those in Table \ref{tab:tab-performance}. The \textit{ARoA-M} model shows the best performance over all the four domains, while it is an interesting observation that the \textit{GR-ADD} model with simple addition operator for Refiner obtains the second best performance. All these prove the importance of the proposed component Refiner in aggregating and selecting the attentive information.

Figure \ref{fig:laptop-performances} illustrates a comparison of four models (\textit{ENCDEC}, \textit{SCLSTM}, \textit{ARoA-M}, and \textit{GR-ADD}) which were trained from scratch on the laptop dataset in a variety of proportion of training data, from $10\%$ to $100\%$. It clearly shows that the BLEU increases while the slot error rate decreases as more training data was provided. Figure \ref{fig:general-models} presents a comparison performance of general models as described in Section \ref{subsec:experimental-setups}. Not surprisingly, the two proposed models still obtain higher the BLEU score, while the \textit{ENCDEC} has difficulties in reducing the ERR score in all cases.
Both the proposed models show their ability to generalize in the unseen domains (TV and Laptop datasets) since they consistently outperform the previous methods no matter how much training data was fed or how training method was used.
These indicate the relevant contribution of the proposed component Refiner to the original AREncDec architecture, in which the Refiner with gating or attention mechanism can effectively aggregate the information before putting them into the RNN decoder. 

Figure \ref{fig:attention} shows a different attention behavior of the proposed models in a sentence. While all the three models could attend the slot tokens and their surrounding words, the \textit{ARoA-C} model with context shows its ability in attending the consecutive words. Table \ref{tab:comparison} shows comparison of responses generated for some DAs between different models. The previous approaches (\textit{ENCDEC}, \textit{HLSTM}) still have missing and misplaced information, whereas the proposed models can generate complete and correct-order sentences.
\section{Conclusion and Future Work}\label{sec:conclusion}
We present an extension of an Attentional RNN Encoder-Decoder model named Encoder-Aggregator-Decoder, in which a Refiner component is introduced to select and aggregate the semantic elements produced by the encoder. We also present several different choices of gating and attention mechanisms which can be effectively applied to the Refiner. The extension, which is easily integrated into an RNN Encoder-Decoder, shows its ability to refine the inputs and control the flow information before putting them into the RNN decoder. We evaluated the proposed model on four domains and compared to the previous generators. The proposed models empirically show consistent improvement over the previous methods in both BLEU and ERR evaluation metrics. In the future, it would be interesting to further investigate hybrid models which integrate gating and attention mechanisms 
in order to leverage the advantages of both mechanisms.


\bibliography{acl2017}
\bibliographystyle{acl_natbib}

\end{document}